\pdfoutput=1

\documentclass[11pt]{article}
\usepackage{ACL2023}

\usepackage{times}
\usepackage{latexsym}

\usepackage[T1]{fontenc}

\usepackage[utf8]{inputenc}

\usepackage{microtype}
\usepackage{cite}
\usepackage{natbib}
\usepackage{times}
\usepackage{latexsym}
\usepackage{CJKutf8}

\usepackage{latexsym}

\usepackage{graphicx}
\usepackage{enumitem}
\usepackage{booktabs}
\usepackage{tabularx}
\usepackage{multirow}
\usepackage{array}

\usepackage{amsmath} 

\usepackage{hyperref}
\newcolumntype{Y}{>{\centering\arraybackslash}X}

\date{}

\begin{document}

\title{RJUA-QA: A Comprehensive QA Dataset for Urology}

\author{
    Shiwei Lyu\textsuperscript{1,*} \,
    Chenfei Chi\textsuperscript{2,*} \, 
    Hongbo Cai\textsuperscript{1} \,
    Lei Shi\textsuperscript{1} \,
    Xiaoyan Yang\textsuperscript{1} \,
    Lei Liu\textsuperscript{1}\\
    \textbf{Xiang Chen\textsuperscript{1} \,
    Deng Zhao\textsuperscript{1} \,
    Zhiqiang Zhang\textsuperscript{1} \, 
    Xianguo Lyu\textsuperscript{2} \,
    Ming Zhang\textsuperscript{2} \, 
    Fangzhou Li\textsuperscript{2} }\\ 
    \textbf{Xiaowei Ma\textsuperscript{2} \,
    Yue Shen\textsuperscript{1,$\dagger$} \,
    Jinjie Gu\textsuperscript{1,$\dagger$}
    Wei Xue\textsuperscript{2,$\dagger$} \, 
    Yiran Huang\textsuperscript{2,$\dagger$}}\\
    \textsuperscript{1}Ant Group\\
    \textsuperscript{2}Department of Urology, Shanghai Jiao Tong University School of Medicine Affiliated Renji Hospital\\
    \texttt{\{lvshiwei.lsw, zhanying, jinjie.gujj\}@antgroup.com}, \\
    \texttt{\{chichenfei, lvxiangguo, zhangming, renjilfz, xuewei, huangyiran\}@renji.com} 
}

\maketitle

\newcommand \footnoteONLYtext[1]
{
	\let \mybackup \thefootnote
	\let \thefootnote \relax
	\footnotetext{#1}
	\let \thefootnote \mybackup
	\let \mybackup \imareallyundefinedcommand
}
\footnoteONLYtext{\textsuperscript{*}These authors contributed equally to this work.} 
\footnoteONLYtext{\textsuperscript{$\ddagger$}Corresponding authors.}

\newpage

\begin{abstract}
We introduce RJUA-QA, a novel medical dataset for question answering (QA) and reasoning with clinical evidence, contributing to bridge the gap between general large language models (LLMs) and medical-specific LLM applications. RJUA-QA is derived from realistic clinical scenarios and aims to facilitate LLMs in generating reliable diagnostic and advice. The dataset contains 2,132 curated Question-Context-Answer pairs, corresponding about 25,000 diagnostic records and clinical cases. The dataset covers 67 common urological disease categories, where the disease coverage exceeds 97.6\% of the population seeking medical services in urology. Each data instance in RJUA-QA comprises: (1) a question mirroring real patient to inquiry about clinical symptoms and medical conditions, (2) a context including comprehensive expert knowledge, serving as a reference for medical examination and diagnosis, (3) a doctor response offering the diagnostic conclusion and suggested examination guidance, (4) a diagnosed clinical disease as the recommended diagnostic outcome, and (5) clinical advice providing recommendations for medical examination. RJUA-QA is the first medical QA dataset for clinical reasoning over the patient inquiries, where expert-level knowledge and experience are required for yielding diagnostic conclusions and medical examination advice. A comprehensive evaluation is conducted to evaluate the performance of both medical-specific and general LLMs on the RJUA-QA dataset. Our data is are publicly available at \href{https://github.com/alipay/RJU_Ant_QA}{https://github.com/alipay/RJU\_Ant\_QA}.

\end{abstract}

\section{Introduction}
Nowadays, online medical diagnosis have become the preferred choice for patients seeking convenient and efficient medical services~\citep{arora2023promise}. Consequently, there has been a notable surge in patients' demands for online medical consultations and inquiries, supported by advancements in internet-based healthcare tools~\citep{singhal2023expertlevel}. Under this background, the explosive development of large language models (LLMs)~\citep{openai2023gpt4,chatgpt} has profoundly facilitated the improvement and application of AI-driven medical technologies within the relevant clinical healthcare scenarios~\citep{jin-etal-2019-pubmedqa}. Leveraging their powerful learning capability for human-machine interaction and modeling complex knowledge, LLMs have demonstrated significant potential to work as intelligent medical assistants in real-world applications.

For one clinical session of a medical consultation, the query of a patient generally contains complicated personal context information, which requires LLMs to recognize and understand the important medical-specific information~\citep{peng2023study}, \textit{e.g.}, the patient's basic information and needs. Then LLMs should be able to work like an experienced clinical expert with the rich medical knowledge, which helps to provide professional and detailed diagnosis and treatment advice via multi-turn dialogues.

However, LLMs still face numerous challenges when dealing with the above-mentioned patients' consultations~\citep{nori2023capabilities,liu2023thinkinmemory}. In detail, existing LLMs usually fail to handling various medical consultations due to a lack of sufficient domain knowledge~\citep{li2023beginner,kamble2023palmyra}, leading to wrong diagnosis and treatment conclusions or irrelevant responses. Moreover, due to the hallucination issue~\citep{hallu_detect,hallu_survey} and weak reasoning ability~\citep{singhal2022large,liévin2023large}, it is greatly difficult to achieve better controllability and accuracy for LLMs when deploying them into the realistic clinical environment. More critically, it is noticed that there exists a shortage of high-quality Chinese medical specialty datasets in the current research landscape. Indeed, the above-mentioned issues pose significant challenges for the applications of LLMs in the medical field.

\begin{figure*}[!t] 
\centering 
\includegraphics[width=1\textwidth]{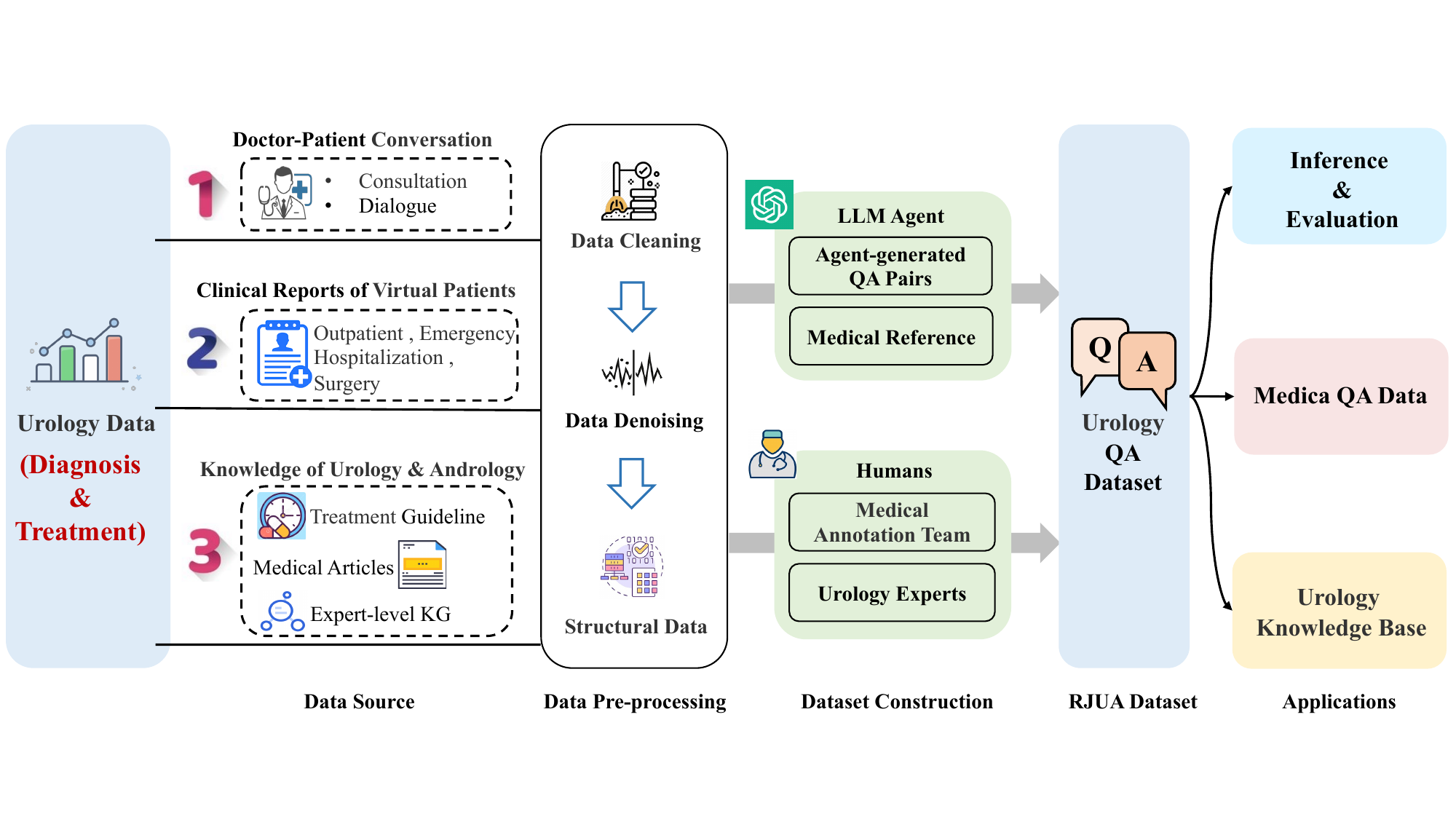} %
\caption{The data construction pipeline of the RJUA-QA dataset.} 
\label{fig:framework}
\end{figure*}

To overcome these challenges, we aim at constructing a high-quality and comprehensive medical specialty QA dataset which (1) has patient consultation simulations with expert-level annotations and (2) requires medical reasoning over the query contexts and clinical knowledge to answer the questions. The data sources mainly involves the virtual patient information derived from the realistic diagnosing cases and clinical experiences of medical experts. An example is shown in Table \ref{dataset_instance}. Each data instance in RJUA-QA comprises: (1) a question mirroring real patient to inquiry about clinical symptoms and medical advice, (2) a context including comprehensive expert knowledge, serving as a reference for medical examination and diagnosis, (3) a response offering the diagnostic conclusion and examination advice, (4) a diagnosed clinical disease as the diagnostic ground-truth, and (5) clinical advice providing recommendations for medical examination. The dataset construction pipeline is illustrated in Figure \ref{fig:framework}.

To our knowledge, RJUA-QA is the first Chinese QA dataset to combine clinical experience with virtual patient query for medical specialty diagnosis and examination advice. Natural language understanding and clinical medical reasoning are required for yielding diagnostic conclusions and examination advice. Furthermore, RJUA-QA provides a medical QA benchmark with the standard evaluation protocols to improve and evaluate the medical reasoning capabilities of LLMs. 


\section{RJUA-QA Dataset}
In this section, we will introduce the statistic information, the dataset characteristics and the data collection pipeline, respectively.
\begin{figure*}[!t] 
\centering 
\includegraphics[width=1.0\textwidth]{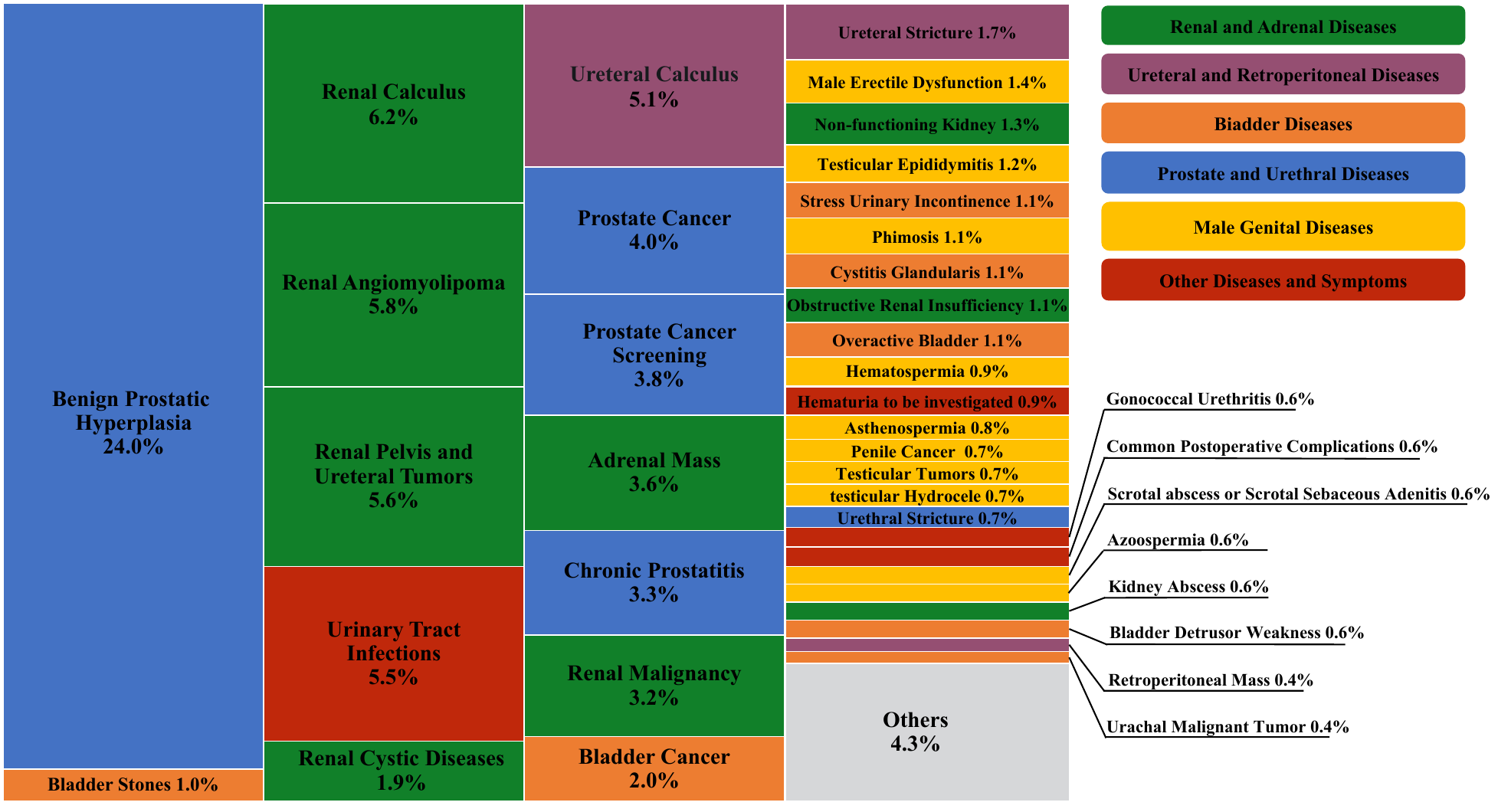} %
\caption{Distribution of Disease Categories in the RJUA-QA Datasets.} 
\label{fig:disease_proportion}
\end{figure*}

\subsection{Data Statistics}
As shown in Table~\ref{tab:medical_terms}, the RJUA-QA dataset contains 2,132 curated Question-Context-Answer pairs, corresponding about 25,000 diagnostic records and clinical cases. Besides, the dataset covers 67 common urological disease categories, where the disease coverage exceeds 97.6\% of the population seeking medical services in urology.

During data selection, according to incidence rates of each disease as well as clinical findings and management, we manually control the occurrence proportion for various diseases in the dataset. The detailed information can refer to Figure~\ref{fig:disease_proportion}. Besides, as one of the most important characteristics of RJUA-QA, the data collection refers to the fact that real patients may perform the diverse subjective descriptions for the same disease, which more authentically replicates the actual diagnostic and treatment scenarios faced by urology specialists.

\begin{figure}[!t] 
\centering 
\includegraphics[width=0.5\textwidth]{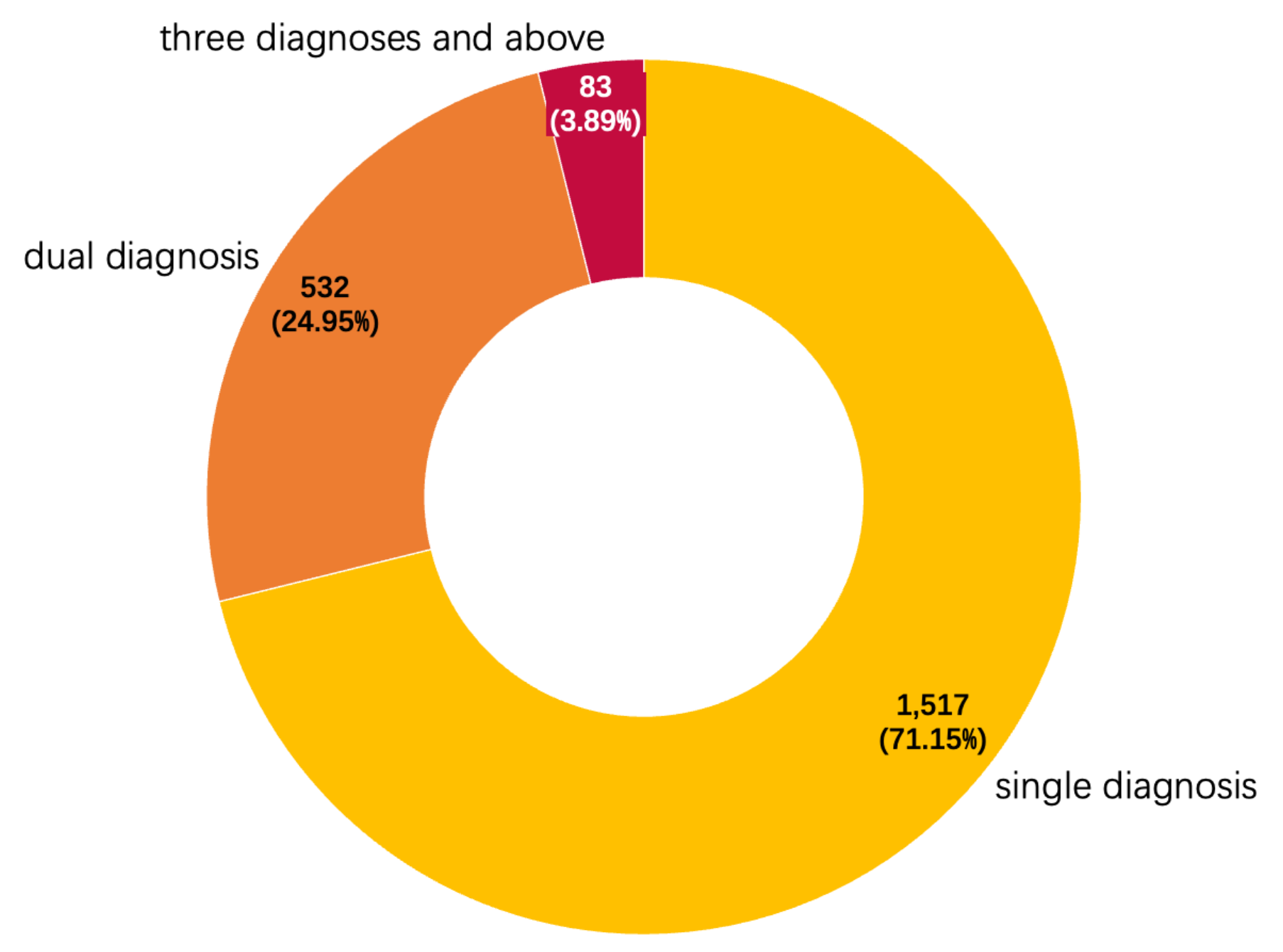} %
\caption{Proportional Breakdown of Urological Disease Diagnosis Categories Across QA Entries.} 
\label{fig:diagnosis_categories}
\end{figure}

Considering the common and prevalent diseases in real clinical patients, including complications caused by primary diseases as well as comorbidities, more than 80\% of the patients in this dataset have multiple kinds of diseases. To reasonably decrease the difficulty of specialist diagnosis, most non-urological comorbidities are directly provided in the questions. As depicted in Figure~\ref{fig:diagnosis_categories}, there are 24.95\% (532/2132) of patients with two urological diagnoses of urology. 3.99\% (83/2132) of patients have three or more urological diagnoses. These patients often require the judgment regarding the primary and secondary diseases or the causal relationship among these diseases. Then the comprehensive diagnostic and examination advice is required to be provided. 

This dataset also provides the reasoning context as reference, sourced from the ``Chinese Urology and Andrology Disease Diagnosis and Treatment Guidelines (2022 Edition)'', major urology textbooks, professional literature from PubMed~\citep{canese2013pubmed}, and the clinicians' experience (more than 10 years).

\subsection{Dataset Characteristics}
\paragraph{Realistic Clinical Background:} The clinical data for the virtual patients is derived based the realistic clinical background, including outpatient diagnosis and treatment, emergency, and inpatient surgical procedures, offering high practical significance and application value.

    
\paragraph{Higher Medical Diversity:} The questions cover multiple organs, sub-specialties, and diseases within urology, with disease coverage accounting for over 95\% of urology patient visits, which helps to enhance the generalizability of the model's application.
    
\paragraph{Interpretability:} The dataset provides detailed and authoritative specialist evidence. This evidence, along with reasoning processes, aids in analyzing the model's reasoning logic and enhances clinical interpretability.
 
\paragraph{Accurate and Rigorous:} The overall dataset is aligned with standard clinical practice, involving the following aspects: urgency and severity of diseases, diagnostic logic, as well as examination and treatment principles. The dataset can enhance the capability to accurately identify the primary disease for patients with multiple diseases. Thus the dataset can well evaluate whether the LLMs can provide professional medical diagnosis and advice.

\subsection{Construction Pipeline}

\subsubsection{Data Source}
\begin{figure*}[!t] 
\centering 
\includegraphics[width=1\textwidth]{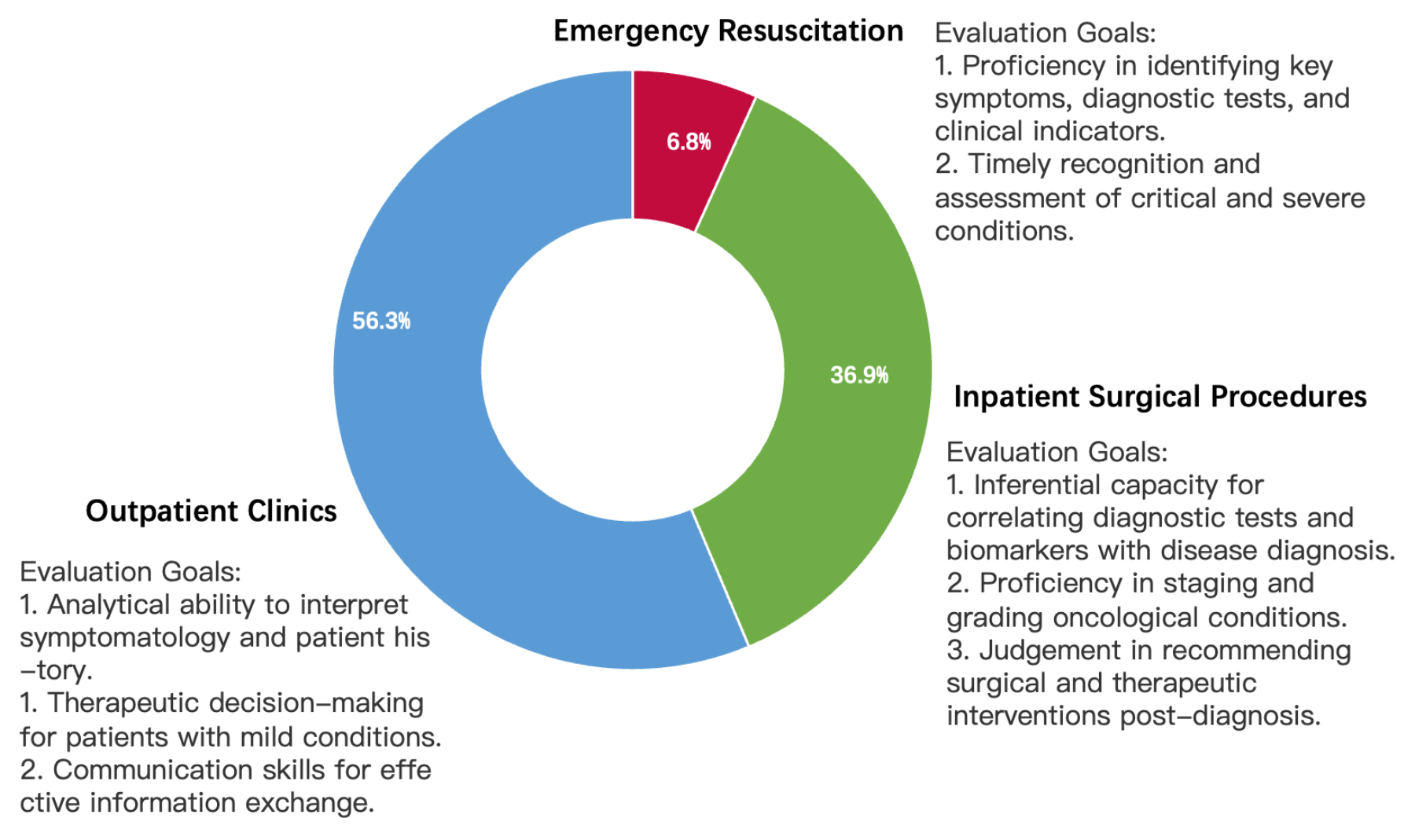} %
\caption{Source and Test Objectives of the RJUA-QA Datasets.} 
\label{fig:source_stat1}
\end{figure*}

Our dataset was developed in collaboration with department of urology Shanghai Renji Hospital. Leveraging their clinical expertise and the powerful generative capabilities of LLM, we created synthetic patient data that accurately reflects real clinical scenarios. The dataset is characterized by the authenticity, precision, and reliability of specialized medical data within the healthcare domain. The synthetic patient data encompassing a wide array of sources, including outpatient diagnoses and treatments, emergency, inpatient surgeries, and procedures, as well as routine public health education. This comprehensive coverage facilitates the evaluation of various clinical application scenarios, as demonstrated in Figure~\ref{fig:source_stat1}. 

\begin{figure*}[!t] 
\centering 
\includegraphics[width=1\textwidth]{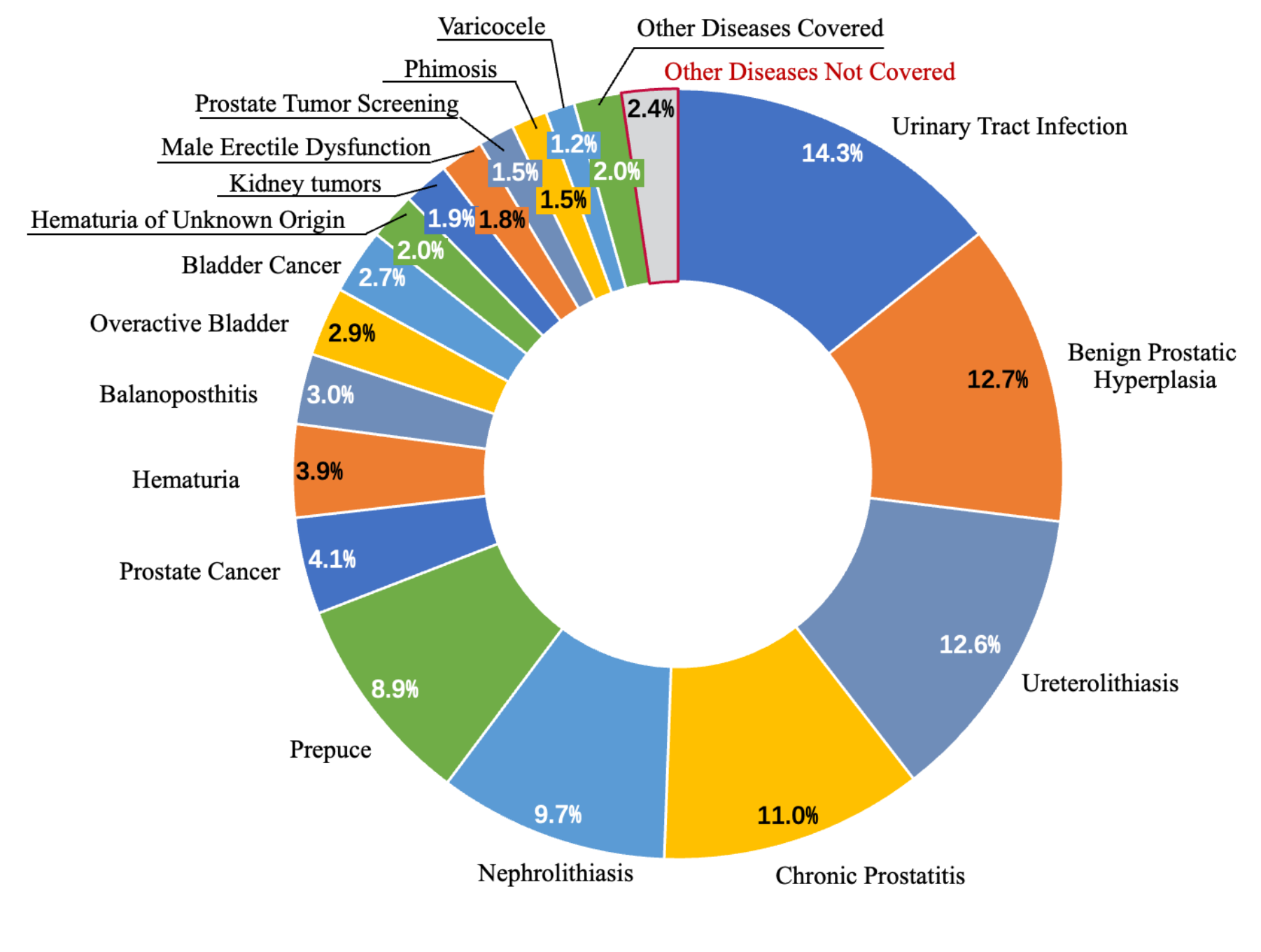} %
\caption{Proportion of Primary Diagnoses in department of urology Shanghai Renji Hospital (2019-2023).} 
\label{fig:source_stat2}
\end{figure*}

Our dataset encompasses a spectrum of urological conditions, covering 10 sub-specialties: urologic oncology, urinary calculi, benign prostatic hyperplasia, male reproductive health, urinary incontinence, reconstructive urology, pediatric urology, and renal transplantation. This comprehensive dataset accounts for 97.6\% of the patient profiles encountered in urological practice, as depicted in Figure~\ref{fig:source_stat2}, ensuring extensive representativeness for research applications.

\subsubsection{Data Pre-processing}
The clinical data of synthetic patients underwent a pre-processing procedure to ensure the high quality and usability, including data cleaning, denoising, and extraction. The overall procedure is described in the follows:

\paragraph{Data Cleaning:} This involves removing any irrelevant or redundant information from the dataset. This step may include correcting spelling mistakes, standardizing date formats, removing duplicates, and dealing with missing or incomplete data entries.

\paragraph{Data Denoising:} The primary aim is to identify and remove any noise present in the data that could potentially distort the analysis. This noise may originate from various sources, including errors in data collection, transmission, or processing. Approaches such as filtering, outlier detection, and statistical methods are employed to smooth the data.

\paragraph{Structured Data Extraction:} This phase is dedicated to the systematic organization and transformation of data into a format amenable to analysis or model development. This process may encompass the parsing of textual data to extricate pertinent fields, the transmutation of unstructured or semi-structured data into a tabular format, and the categorization or encoding of data to simplify subsequent processing steps. The culmination of this phase is the attainment of a streamlined and methodically organized dataset, primed for ensuing stages of data analysis or machine learning endeavors.

\subsubsection{Dataset Construction}
\paragraph{Generate QA Pairs with LLM:} 
 We assign roles to LLM, \textit{e.g.}, GPT-3.5~\citep{brown2020language}, to act as intelligent agents through specific instruction. Building on structured data extracted from medical doctor-patient conversations and clinical reports of virtual patients, we facilitate the generation of corresponding Question-Answer (QA) pairs. This process harnesses the capabilities of advanced LLMs to simulate nuanced interactions in a medical context, providing a novel and practical application of AI in healthcare.

\paragraph{Collect Medical Literature as Reference Context:}

(1) Under the supervision of medical experts, pertinent text fragments from medical guidelines are manually extracted with a focus on broad coverage, serving as potential context candidates.

(2) For each QA (Question-Answer) data entry, we combine it with every context candidate related to its associated disease. Then, a LLM is utilized to assess whether the given context is relevant. Upon receiving a response from the LLM, the context candidates that are identified as matching are then incorporated into the dataset as the context for that specific QA entry.

(3) To enhance the complexity of the task, context candidates unrelated to the disease are also randomly chosen and undergo the same QA-context matching process. The contexts that align with the QA pair are added to the dataset, acting as distractors and thereby increasing the difficulty of the task.

(4) The dataset is subjected to manual verification to ensure accuracy. During this process, any contexts that were initially overlooked are identified and added to the dataset.

\paragraph{Human Based Data Calibration:} Our methodology involved a systematic three-tiered review and validation process for each Q-context-A triad. This process was executed by a medical annotation team with clinical expertise, in conjunction with the urology expert panel from Shanghai Renji Hospital. The review focused on six key dimensions. These included the precision of medical terminology and the coherence between questions and answers. Also assessed were the relevance of the provided context and its role as pivotal evidence. The logical soundness within the answers and the accuracy of the resultant diagnoses were critically evaluated.

\paragraph{Formulate the Structural Data Format:}
Our approach involved the careful curation of QA pairs and logical inference steps into a structured data format, enhanced by the development of custom reasoning evaluation metrics. This carefully assembled dataset fulfills two key objectives. Firstly, it aids in the fine-tuning of Large Language Models (LLMs) to utilize specialized medical knowledge bases, thereby improving diagnostic accuracy. Secondly, it offers a solid framework for assessing the inferential capabilities of LLMs in medical diagnosis. This method sets the stage for advanced AI applications in healthcare, where precision and reliability are crucial.

\section{Experiments}

\begin{table*}[!t]
\centering
\begin{tabular}{cc|c|c}
\toprule
\multicolumn{2}{c|}{\multirow{2}{*}{\textbf{LLM}}} & \multicolumn{2}{c}{Predicted} \\ \cmidrule{3-4}
& & Positive & Negative \\ \cmidrule{1-4}
\multicolumn{1}{c|}{\multirow{2}{*}{GroundTruth}} & Positive & TP & FN \\ \cmidrule{2-4}
\multicolumn{1}{c|}{} & Negative & FP & TN \\ \cmidrule{1-4}
\end{tabular}
\caption{Confusion Matrix. FP, TN, FN, TP are the shorts for False Positive, True Negative, False Negative, and True Positive, respectively.}
\label{confusion}
\end{table*}

\begin{table*}[!t]
\centering
\begin{tabular}{l|c|ccc}
\toprule
\multirow{2}{*}{\textbf{LLM}} & \multirow{2}{*}{\textbf{Rouge-L}} & \multicolumn{3}{c}{\textbf{F1}}                                                   \\ \cmidrule{3-5} 
                     &                          & \multicolumn{1}{l|}{disease}  & \multicolumn{1}{l|}{advice}   & total    \\ \midrule
Huatuo-13B           & 0.183456                 & \multicolumn{1}{l|}{0.496833} & \multicolumn{1}{l|}{0.042711} & 0.345459 \\ \midrule
GPT-3.5              & 0.233997                 & \multicolumn{1}{l|}{0.458559} & \multicolumn{1}{l|}{0.019254} & 0.312124 \\ \midrule
Baichuan-13B         & 0.204206                 & \multicolumn{1}{l|}{0.486466} & \multicolumn{1}{l|}{0.030048} & 0.334327 \\ \midrule
ChatGLM3-6B          & 0.212661                 & \multicolumn{1}{l|}{0.500212} & \multicolumn{1}{l|}{0.038699} & 0.346374 \\ \midrule
Qwen-7B              & 0.081869                 & \multicolumn{1}{l|}{0.403177} & \multicolumn{1}{l|}{0.053067} & 0.286474 \\ \midrule
\end{tabular}
\caption{The evaluation results for general and medical-specific LLMs on the RJUA-QA dataset.}
\label{results_llm}
\end{table*}

\subsection{Baseline Setup}
\textbf{Huatuo GPT.} HuatuoGPT~\citep{zhang2023huatuogpt} is a domain-specific LLM for medical consultation. HuatuoGPT leverages both distilled datavfrom ChatGPT and real-world data from doctors in the supervised fine-tuned stage, which trains a reward model to align the language model with the merits following an reinforced learning from AI feedback.

\noindent \textbf{GPT-3.5.} GPT-3.5 is an advanced language model developed by OpenAI. One of the key features of GPT-3.5 is its ability to perform a wide range of natural language processing tasks, such as language translation, summarization, question answering, and text completion. It can generate responses that are contextually relevant and coherent with the given input.

\noindent \textbf{Baichuan.} Baichuan~\citep{baichuan2023baichuan2} is an open-source large-scale multilingual language model containing 13 billion parameters, which is trained from scratch on 2.6 trillion tokens. This model excels at dialogue and context understanding.

\noindent \textbf{ChatGLM.} ChatGLM~\citep{zeng2022glm} is an open-source bilingual language model based on the General Language Model (GLM) framework. This model contains 6.2 billion parameters with specific optimization, involves supervised fine-tuning, feedback bootstrap, and reinforcement learning with human feedback. We include ChatGLM3 as a baseline for evaluations.

\noindent \textbf{Qwen.} QWen~\citep{qwen} is a comprehensive language model series that encompasses distinct models with varying parameter counts. The base language models
consistently demonstrate superior performance across a multitude of downstream tasks.

\subsection{Evaluation Protocols}
The dataset is designed to enhance the capabilities of large language models in medical logical reasoning and serve as an evaluation benchmark for applications in critical and controllable scenarios. The evaluation scheme assesses the model's responses from two perspectives:

\paragraph{Diagnosis and Advice Accuracy:} The F1 score is utilized to measure the accuracy for LLMs' diagnosis and treatment. According to Table \ref{confusion}, F1 score is is formulated as:
\begin{equation}
    \text{F1} = 2\times \frac{P\times R}{P + R},
\end{equation}
where $P=\frac{\text{TP}}{TP+FP}$ denotes the precision and $R=\frac{TP}{TP+FN}$ denotes the recall. A weighted sum of F1 score for diagnosis and advice is adopted to obtain the final accuracy, \textit{i.e.}, 2/3 for diagnosis and 1/3 for advice.

\paragraph{Overall Response Quality:} To evaluate the overall quality of the LLMs' responses, Rouge-L is exploited to calculate the longest common sub-sequence (LCS) between the generation and reference. LCS is the sequence of words that appear in the same order in both summaries with the maximum length. Rouge-L then computes precision, recall, and F1 score based on the LCS.

\subsection{Main Results}
As shown in Table \ref{results_llm}, GPT-3.5 exhibits the highest Rouge-L score. The main reason is that GPT-3.5 can generate more human-like sentences, benefiting from its larger model parameters and better language abilities. ChatGLM3 and Qianwen could obtain the best performance for disease diagnosis and treatment advice, possibly because these models encode more medical knowledge during pre-training (especially academic vocabulary). In addition, Qianwen achieves a lower Rouge-L score, since its generated sentences are too long, resulting in lower accuracy. 

\section{Conclusion}
In this paper, we introduced a novel medical specialty QA dataset called RJUA-QA, which facilitate machine intelligence in producing precise diagnostic outcomes. RJUA-QA is the first QA dataset for clinical medical reasoning, requiring expert knowledge and experience in yielding diagnostic conclusions and examination guidance.

There are several featured characteristics of RJUA-QA, \textit{i.e.}: (1) The synthetic patient data is derived from the realistic clinical background. (2) The questions cover various urological organs, sub-specialties, and diseases, exhibiting higher diversity. (3) The dataset offers detailed medical evidence for reasoning with explicit reasoning interpretability. (4) Data quality is checked by clinical expert with accurate diagnostic results and scientific examination principles.

We provide a detailed description for the data collect, data characteristics, and statistical analysis. We will continually optimize the benchmark, providing strong supports for research and application of artificial intelligence in the medical field.

\section{More Discussion}
In the future, our team plans to continue iterating and optimizing the RJUA-QA Datasets, including incorporating more real-world clinical experience data, increasing coverage of rare and uncommon diseases in the disease database, and enriching more medical scenarios, dialogue methods, and emotional appeals. Additionally, we will also develop multi-turn QA datasets that are more aligned with the actual multi-turn dialogue scenarios in medical consultations. This will provide researchers with more diverse and challenging data resources. Furthermore, we will focus on evaluating benchmarks for large models in terms of reasoning ability and practical application in medical scenarios. We will explore new methods and technologies to improve the deployment capabilities of models in serious and controlled environments.

We hope to contribute to the research and application of artificial intelligence in the medical field through continuous efforts. We aim to promote the development of intelligent medical assistants to better serve patients and healthcare professionals, which can improve the quality and efficiency of healthcare services. 



\bibliography{acl_refer}
\bibliographystyle{acl_natbib}


\begin{table*}[!t]
\centering
\begin{tabular}{c|p{10cm}}
\toprule
\textbf{Question} & \begin{CJK}{UTF8}{gbsn}医生你好，我是一名71岁的男性患者。最近我晚上经常起夜去上厕所，每晚大约要起3-5次，而且尿流感觉变得越来越细。我还发现射精变得困难，有时候还会出现射精疼痛的情况。不过我没有尿血或尿急。请问这是什么原因引起的呢？我应该如何治疗？\end{CJK}\\
\midrule
\textbf{Context} & 
    \begin{CJK}{UTF8}{gbsn}临床诊断：LUTS包括频繁夜尿、排尿困难以及尿频尿急。随着患者的年龄增加，尿急、尿急且夜尿等情况将会增多;排尿困难包括排尿迟缓、排尿困难以及排尿间断等;排尿困难在排尿终末不尽感、尿后滴沥等。有关LUTS的中老年男性更容易发生勃起功能障碍(erectile dysfunction，ED)[5],且ED与LUTS的严重程度相关性6-7。
    临床诊断1：【2 诊断与评估】勃起障碍的检查包括：病史询问、国际前列腺症状评分(International Prostate Symptom Score, IPSS)、体格检查、尿常规检查、血清前列腺特异性抗原(prostate specific antigen,PSA)检测、经直肠数字直肠检查等。\end{CJK} \\
\midrule
\textbf{Answer} & \begin{CJK}{UTF8}{gbsn}您好，根据您的症状描述和年龄，很可能是前列腺增生引起的。前列腺增生是一种常见的男性疾病，随着年龄的增长，前列腺会逐渐增大，压迫尿道，导致尿流减细、尿线变弱以及尿不尽等症状。根据您的情况，我们建议您进行前列腺。前列腺特异性抗原（PSA）和前列腺系统B超检查，治疗方面，可以根据您的具体情况选择合适的治疗方法，包括药物治疗、物理治疗或手术治疗。建议您尽快到医院就诊，进行详细的检查和评估，以便制定最合适的治疗方案。\end{CJK}\\
\midrule
\textbf{Disease} &\begin{CJK}{UTF8}{gbsn}前列腺增生\end{CJK} \\
\midrule
\textbf{Advice} &\begin{CJK}{UTF8}{gbsn}体检项目，前列腺特异性抗原（PSA），泌尿系统B超\end{CJK} \\
\bottomrule
\end{tabular}
\caption{An instance of the RJUA-QA dataset.}
\label{dataset_instance}
\end{table*}

\begin{table*}[!h]
  \centering
  \small
  \begin{tabularx}{\textwidth}{YYYYYY}
    \toprule
    \textbf{Renal and Adrenal Diseases} & \textbf{Bladder Diseases} & \textbf{Prostate and Urethral Diseases} & \textbf{Male Genital Diseases} & \textbf{Ureteral and Retroperitoneal Diseases} & \textbf{Other Diseases and Symptoms} \\
    \midrule

    Adrenal Mass & Bladder Cancer & Benign Prostatic Hyperplasia & Balanoposthitis &Ureteral Calculus & Urinary Tract Infections \\
    \midrule
    Renal Malignancy & Overactive Bladder & Acute Prostatitis & Phimosis & Ureteral Stricture & Common Postoperative Complications\\
    \midrule
    Renal Angiomyolipoma & Bladder Stones & Chronic Prostatitis &  Prepuce & Retroperitoneal Mass & Hematuria of Unknown Origin \\
    \midrule
    Renal Pelvis and Ureteral Tumors & Bladder Diverticulum & Prostate Cancer & Preputial Scars & Pelvic Lipomatosis &  \\
    \midrule
    Renal Calculus & Bladder Foreign Body & Prostate Cancer Screening & Frenular Tear & Retroperitoneal Fibrosis & \\
    \midrule
    Renal Cystic Diseases & Vesicovaginal Fistula & Urethral Stricture & Penile Cancer \\
    \midrule
    Renal Abscess & Vesicorectal Fistula & Urethral Diverticulum & Scrotal Gangrene \\
    \midrule
    Renal Trauma & Cystitis Glandularis & Urethral Foreign Bodies & Scrotal Abscess \\
    \midrule
    Spontaneous Renal Rupture & Bladder Detrusor Weakness & Gonococcal Urethritis & Scrotal Trauma \\
    \midrule
    Obstructive Nephropathy & Stress Urinary Incontinence & & Seminal Vesiculitis \\
    \midrule
    Non-functioning Kidney & Neurogenic Bladder & & Spermatic Cord Cyst \\
    \midrule
    Duplex Kidney & Urachal Cancer & & Epididymal Cyst \\
    \midrule
    Polycystic Kidney Disease & Urachal Cyst & & Testicular Tumor \\
    \midrule
    Renal Transplantation & Urachal Anomaly & & Epididymo-orchitis \\
    \midrule
    & & & Testicular Torsion \\
    \midrule
     &  & & Hydrocele Testis \\
     \midrule
     &  & & Cryptorchidism \\
     \midrule
    &  & & Erectile Dysfunction \\
    \midrule
    &  & & Male Infertility \\
    \midrule
     & & & Azoospermia \\
     \midrule
    & & & Oligozoospermia \\
    \midrule
    & & & Hematospermia \\
    \bottomrule
  \end{tabularx}
  \caption{Inventory of Diseases in the RJUA-QA Datasets}
  \label{tab:medical_terms}
\end{table*}

\end{document}